# A Textural Approach to Palmprint Identification

Mrs. Rachita Misra
Prof. & Head Department of IT
CV Raman College of Engineering
Bhubaneswar, Orissa, India.

Mrs. Kasturika B. Ray
Institute of Technical Education & Research
Siksha 'O' Anusandhan University, Bhubaneswar
(phd. contd.), Orissa, India.
Mrs Rachita Misra

***Abstract*** – Biometrics which use of human physiological characteristics for identifying an individual is now a widespread method of identification and authentication. Biometric identification is a technology which uses several image processing techniques and describes the general procedure for identification and verification using feature extraction, storage and matching from the digitized image of biometric characters such as Finger Print, Face, Iris or Palm Print. The current paper uses palm print biometrics. Here we have presented an identification approach using textural properties of palm print images. The elegance of the method is that the conventional edge detection technique is extended to suitably describe the texture features. In this technique all the characteristics of the palm such as principal lines, edges and wrinkles are considered with equal importance.

## I. Introduction

To be able to use a human physiological and behavioral characteristic as a biometric identifier, it must satisfy the characteristics of universality, distinctiveness and permanence. Also it should be easy to collect, store and process to provides reasonable performance. Several types of biometric systems have been employed in different domains that require some sort of user verification. Biometric identification can be considered as the technology that describes procedure for identification and verification using feature extraction, storage and matching from the digitized image of biometric characters such as Finger Print, Face, Iris or Palm Print.

Palm print can be characterized by the geometry of the Heart, Head and Life Lines and the presence of several wrinkles and ridges or crease in the palm as can be seen in Fig 1.

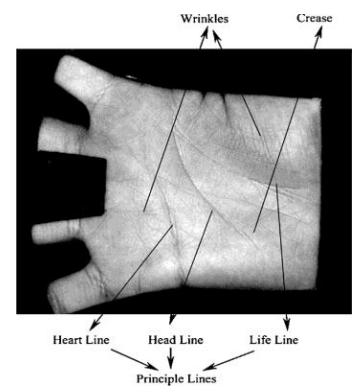

**Fig. 1:** Palmprint image with the principal lines, wrinkles and ridges or crease.

These principal lines and the end points of principal lines (Datum Points) have been used to extract useful palm print features for identification purpose by Zhang et al. [1]. Matching of the Line feature geometry has been found easy for computation, and reported to be powerful for its tolerance to noise and high accuracy in palm print verification. Palm print alignment and classification by using invariant geometrical features has been reported by Wenxin et al [2]. The principal components analysis (PCA) and verification on a database images strongly confirm the robustness of calculating a comprehensive set of selected hand geometry features [3]. Guangming et al. have suggested use of KLT and extraction of Eigen vectors to reduce the dimension of feature space and provides a efficient method of recognition[4].

Several different methods and issues relating to image acquisition, feature extractions and classification and identification have been addressed by several researchers [5, 10]. Issues relating to security and privacy that might enforce accountability and acceptability standards have been discussed Prabhakar et al. [6]. Projecting palm prints from a high-dimensional original palm print space to a significantly lower dimensional feature space using fisher palms have been proposed to efficiently discriminate different palms by Xiangqian et al. [7]. Another innovative method given by Wai Kin Kong uses 2-D Gabor filters to extract palm print texture features [8]. A method of locating and segmenting original palm print space to region of interest (ROI) using elliptical half-rings has been reported to improve the identification by Poon et al. [9]. Palm-print features have been extracted from the ROI by using Sobel and morphological operations [10].

An analysis and comparison of Geometric, Statistical and Textural methods available in the literature was presented by us [11]. In the current paper we describe a generalized Palm Print Biometric system and suggest a simple texture feature based method for palm print identification with encouraging results.

## II. Palm Print Acquisition and Processing

The general steps in the palm print biometric system are -

• Capturing Palm Prints image database for selected users.

• Extraction of features for each class of palm prints and creation of the classified feature database.

• Extraction of features for the input image .

• Finally, the input image features are matched with the stored feature database and the class with the highest matching score is identified as output.

Palm print can be extracted from Hand images of every user. Palm print image can also be captured by using a scanner and digitized. The digitized palm print images are stored in a computer database. Some pre-processing may be necessary to bring all the palm print images to a common coordinate system. Pre-processing techniques may be necessary to improve the quality of the images. Features extraction techniques are applied on the palm print images. The feature database is classified and indexed as several images may belong to the same person.

A distance measure is used to measure the similarity between the input image features and the palm print classes in the data base. The challenge is to use an appropriate feature set which represents the palm and can be used to classify the palm print image data base. The choice of similarity measure is important to be able to assign the correct class to the input image.

### III.Proposed Method

Palm print identification methods have used features extracted from the principal lines or the wrinkles. Principal lines features can have similarity across different palms. Wrinkles are important characteristics but it is difficult to extract them accurately. The proposed method uses the *Edge Features* of the palm to provide a description of texture features.

Edge detection using masks has been widely used in image processing literature. The number of edges in a region provides a measure of signal "busyness" or "edgyness" in that area. A palm print image can be divided into several areas and the number of edges over these areas can be used to define a feature vector for the image.

For an illustration the current method uses four equal regions denoted as *LT* (left top), *LB* (left bottom), *RT* (right top) and *RB* (right bottom) and the number of edges for each region is used to provide a feature vector for the image. A set of such feature vectors can be stored by taking several samples of the same palm. A database of known palmprints' feature vectors are then stored as classified training feature database.

When an unidentified palm is presented then its "edgyness" feature is extracted for the four regions giving an unknown (test) feature vector. The matching of feature between the unknown *(U)* palm feature vector and the database of training *(T)* vectors will identify the palmprint to one set in the training data base.

The *city block distance* is used to measure the similarity of two palm print feature vectors.

$$D\ (U,\ T) = |LTeu\text{-}LTet| + |LBeu\text{-}LBet| + |RTeu\text{-}RTet| + |RBeu\text{-}RBet|$$

Where *U* denotes unknown palm feature vector, *T* denotes a training palm feature vector, the subscript '*et*' denotes training palm "edgyness" and '*eu*' denotes unknown palm "edgyness", and *LT, LB, RT* and *RB* denote the four regions.

The distance of the single unknown palm to a set of samples for a known classified palm in the training database is obtained by averaging the distances using,

$$D\,i = \sum_{J=1}^{N} D\,(\,U\,,\,T\,i\,,j\,)\ \ /\ \ \sum^{N} j$$

j=1

Where i denotes the unknown class *( i= 1, M )* and j denotes the individual samples in the training class, N being the total no. of sample in a class.

The process can be extended to dividing the image into 8 or 16 equal regions. The edge feature in this case is an 8-component or a 16-component vector.

The different steps in the proposed methods have been illustrated in the system Diagram of Fig 2.

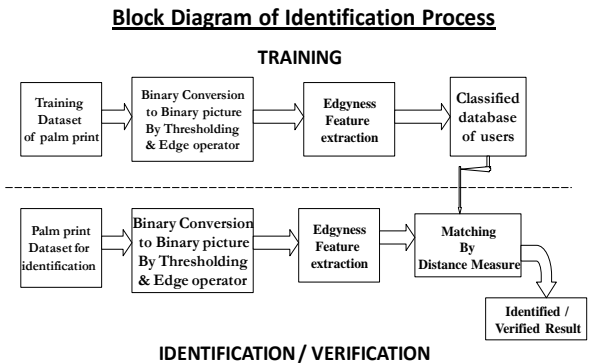

**Block Diagram of Identification Process**

**Fig. 2: Palm Print Biometric System**

**IV. Experimental Result**

**(a) Palm print Acquisition:-**

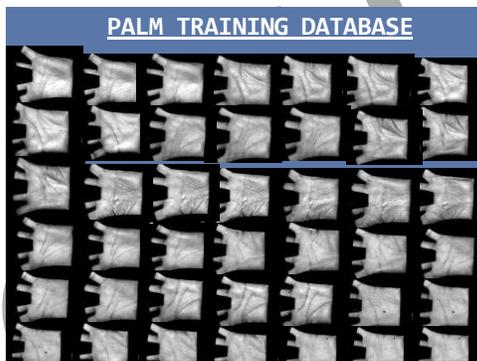

**Fig. 3: Snapshot of the Database**

Palm print images have been collected from internet (total 600 peg free poly u database collected from Hong Kong polytechnic university). The image size is 384x284 pixels in 256 gray levels. The entire palm was preserved, fingers and thumb were omitted. The database consists of palm print of 60 different individuals. Each data set has 12 samples of left palm and 12 samples of right palms.

**(b) Preprocessing:-**

For the experiment 10 classes of left palm print having 12 samples each was considered. For each set, 6 samples were taken as training samples and 6 samples for test. Initially each palm is divided to 4 equal regions denoted as *LT* (left top), *LB* (left bottom), *RT* (right top) *and RB* (right bottom).

**(c) Texture Feature Extraction:-**

Experiments have been conducted to select a suitable edge detector for the palm print Texture Feature using *threshold log,*

*Laplacian and Sobel* operator over a *3X3* area (Fig 4). Further an *8-connectivity* region is used to filter out unwanted edges.

The number of connected lines provides the measure of "edgyness". The feature vector of each palm describing the texture pattern thus consists of the "edgyness" value for the four regions of palm as extracted above (Fig 5, Fig 6.a to 6.d).

Thus for each palm sample in the training database a 4-element feature vector containing the "edgyness" of each region was stored as training feature database.

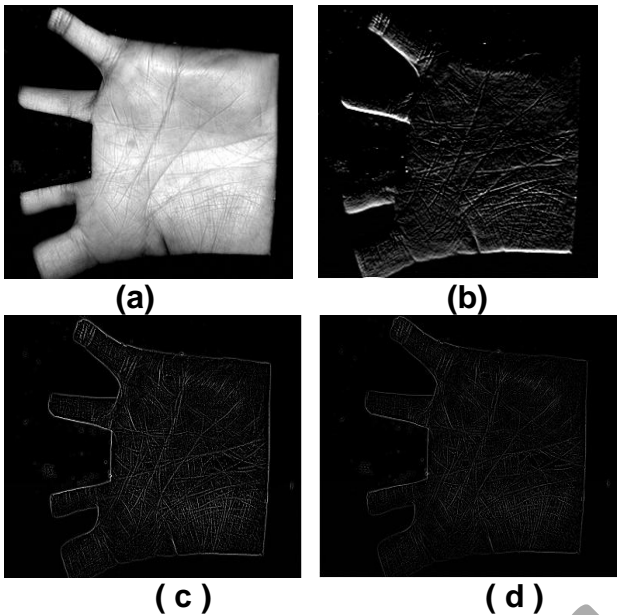

**Fig. 4: (a) Original image, (b) Sobel Threshold image, (c) Log Threshold , (d) Laplacian Threshold.**

The training database thus contains *MxN* feature vectors where M is the number of palm classes *(M=10)* and N is the number of training samples *(N=6)* in each class.

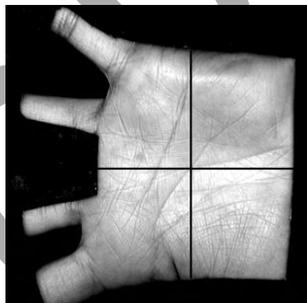

**Fig. 5: Original image with 4 regions.**

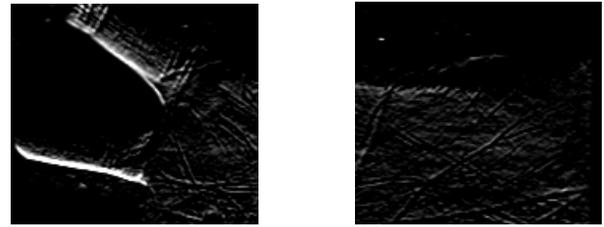

**Fig. 6.a: Extracting edgyness of Left Top(LT) and of Right Top(RT)**

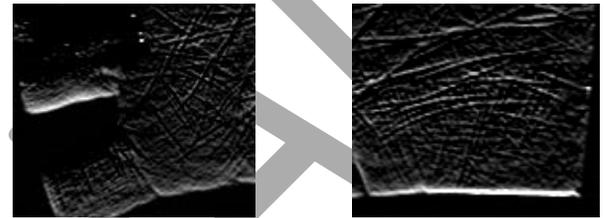

**Fig. 6.b: Extracting edgyness of Left Bottom(LB) and of Right Bottom(RB)**

**(d) Test Data Identification :-**

Palm prints for the test (unknown) samples were selected randomly from the test database. The test palm was divided into four equal regions as given in (b) and texture feature vector was extracted as described in ( c ).

The identification problem is now to classify the test palm to one of the 10 sets of palm in the training database, by comparing the feature vector of the test palm with the feature vector database of training samples. This is achieved by :

- Determining the average distance of unknown sample from the *N* samples of each of the *M* classes *(here N=6, M= 10).*

- The minimum average distance identifies the palm class to which the unknown sample belongs.

The method of pre-processing, feature extraction and test data identification was then repeated with 8 and 16 equal regions.

### (e) Result Analysis :-

All the experiments have been conducted using Matlab. The experimentations with different *edge gradient operator*s showed best result for the *Sobel operator* (Fig 4.b).

To determine the effectiveness of the proposed method we need to examine the correct identification rate *(R) . R* can be defined as

$R = $ <u>*No of test samples correctly classified*</u>
   *Total number of test samples selected*

The correct identification rate using the averaging distance method to a class of palm prints with 4, 8 and 16 regions was found to be 90%, 70% and 80% respectively.

The process of finding minimum distance between known and training samples was then iterated over the individual members of the training classes. However, instead of using all the *M x N* samples of the entire test database only the two test classes which have least average distance from the unknown image feature vector were chosen. This reduces the number of operations required for identification. In this second iteration the correct detection rate was found to be 100% for 4, 8 and 16 regions.

### V. Discussion and Conclusion

This paper describes a new approach to palm print identification using texture feature extraction. Simple edge processing has been used to describe the texture. Palm print identification involves the search for the best matched test samples with the input palm print in the texture feature space. The feature vector here consists of count of connected edges. The correct detection rate with a single iteration is between 70-90%, where as in the second iteration it is found to be100%.

The proposed texture detection method combines the wrinkles, ridges and lines characteristics available in the palm print. The major advantage of this method is its simplicity of implementation and the small size of feature vector. Comparison of identification rate with other methods reported in the literature shows comparable or lower correct detection rate [1, 4, 8, 9] considering the two iterations. Additional improvements in the first iteration results can be achieved by extracting a region of interest for each palm before the feature vector extraction. This will involve some additional pre-processing.

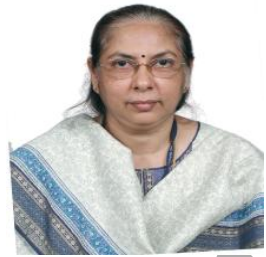

**Dr. Rachita Misra** has a Post graduate degree in Mathematics and Ph.d in the field of Digital Image processing. She has around 25 years of industrial experience in Information Technology solutions and consultancy. She has nearly 10 years of research / teaching experience. She has several publication in the areas of Image Processing, Data Mining and Software Engineering in international and national journals, seminars and conferences.

She is currently heading the Information Technology Department of C.V.Raman College of Engineering, Bhubaneswar, India. She is the editor of International Journal of Image Processing and Vision Science and technical reviewer of several international and national conferences.  She is life member of Computer Society of India (CSI), Indian unit of Pattern Recognition and Artificial Intelligence (IUPRAI) , Indian Science Congress Association (ISCA) and Odisha Information Technology Society (OITS).

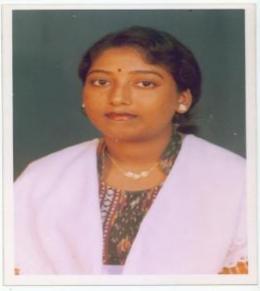

**Kasturika B. Ray** received her M.I.T (Master of Information Technology) Post Graduate degree from Manipal Deemed University, Karnataka 2003, and continuing her Ph.D. research in Computer Science and Engineering, SOA University, Bhubaneswar, under the guidance of Dr. Mrs. Rachita Misra. She has published one International Journal research paper and presented in 2 National conferences and has attended 10 National Workshops / Seminars etc. Her area of interest is Digital Image Processing.